\documentclass[preprint,3p]{elsarticle}
\usepackage{framed,multirow}
\usepackage{amsmath,amssymb,amsfonts}
\usepackage{latexsym}
\usepackage{textcomp}
\usepackage{bm,array,booktabs}
\usepackage{graphicx,subfigure}
\usepackage{multirow,tabularx}
\usepackage{xcolor}
\usepackage{tikz}
\usetikzlibrary{shapes,arrows,fit,positioning}
\DeclareMathOperator*{\argmax}{argmax}
\usepackage{lineno,hyperref}

\biboptions{comma,square,sort&compress}
\modulolinenumbers[5]

\journal{Information Sciences}

\begin{document}

\begin{frontmatter}

\title{Stochastic Mutual Information Gradient Estimation for Dimensionality Reduction Networks}

\author[1,2,3]{Ozan~\"{O}zdenizci\corref{cor1}} 
\cortext[cor1]{Corresponding author: O.~\"{O}zdenizci\corref{cor1}}
\ead{oezdenizci@tugraz.at}
\author[1]{Deniz~Erdo\u{g}mu\c{s}}

\address[1]{Department of Electrical and Computer Engineering, Northeastern University, Boston, MA, USA}
\address[2]{Institute of Theoretical Computer Science, Graz University of Technology, Graz, Austria}
\address[3]{TU Graz - SAL Dependable Embedded Systems Lab, Silicon Austria Labs, Graz, Austria}

\begin{abstract}Feature ranking and selection is a widely used approach in various applications of supervised dimensionality reduction in discriminative machine learning. Nevertheless there exists significant evidence on feature ranking and selection algorithms based on any criterion leading to potentially sub-optimal solutions for class separability. In that regard, we introduce emerging information theoretic feature transformation protocols as an end-to-end neural network training approach. We present a dimensionality reduction network (MMINet) training procedure based on the stochastic estimate of the mutual information gradient. The network projects high-dimensional features onto an output feature space where lower dimensional representations of features carry maximum mutual information with their associated class labels. Furthermore, we formulate the training objective to be estimated non-parametrically with no distributional assumptions. We experimentally evaluate our method with applications to high-dimensional biological data sets, and relate it to conventional feature selection algorithms to form a special case of our approach.\end{abstract}


\end{frontmatter}


\section{Introduction}
\label{sec:intro}

In supervised discriminative model learning, given a finite number of training data samples, optimal exploitation of the information content in the extracted features with respect to their class conditions is essential. Applications in various research fields have developed different domain-specific methods for feature learning and subsequent supervised model training \cite{Lemm:2011,Larranaga:2006,Jiang:2016}. Many exploratory applications in practice are further characterized by high-dimensional feature representations where the dimensionality reduction problem is to be addressed.

One traditional approach towards supervised dimensionality reduction is \textit{feature selection}, referring to the process of selecting the most class-informative subset from the high-dimensional feature set and discarding others \cite{Guyon:2003}. Particularly, feature selection based on information theoretic criteria (e.g., maximum mutual information) have shown significant promise in earlier studies \cite{Battiti:1994,Kwak:2002}. Although selecting a class-relevant subset of features leads to intuitively interpretable and preferable learning algorithms, feature ranking and selection algorithms are known to potentially yield sub-optimal solutions due to their inability to thoroughly assess feature dependencies \cite{Erdogmus:2008,Torkkola:2008}. In that regard, \textit{feature transformation} based dimensionality reduction methods provide a more robust alternative \cite{Guyon:2003}, which have been also studied in the form of information theoretic projections or rotations \cite{Torkkola:2003,Hild:2006,Faivishevsky:2012}.

These latter studies constitute the basis of our current work, in which we address the problem of learning feature transformations based on a maximum mutual information criterion between transformed features and their associated class labels using artificial neural networks. Beyond exhaustively aiming to estimate the mutual information quantity between continuous valued features and discrete valued class labels across training data samples \cite{Ross:2014,Gao:2017}, we claim that feature transformations under a maximum mutual information criterion can be obtained by using a stochastic estimate of the gradient of the mutual information. This feature transformation approach can be further realized as a dimensionality reduction neural network which: (1) can be trained via standard gradient descent, (2) reduces the inference time to a single forward pass through the learned network, and (3) simplifies the overall supervised dimensionality reduction problem by alleviating the need for heuristic and sub-optimal feature selection algorithms.

In this paper we present MMINet, a generic dimensionality reduction neural network training procedure based on maximum mutual information criterion between the network-transformed features and their associated class labels. We derive a stochastic estimate of the gradient of the mutual information between the continuous valued projected feature random variables and discrete valued class labels, and use this stochastic quantity for the loss function in artificial neural network learning. Furthermore, we formulate the training objective non-parametrically, relying on non-parametric kernel density estimations to approximate projected feature space class-conditional probability densities. We interpret our approach as determining a manifold on which transformations of the original features carry maximal mutual information with the class labels. Subsequently, feature selection becomes a special sparse solution case of all possible solutions that MMINet can provide when it is restricted to a single linear layer architecture. For our empirical assessments, we demonstrate our results on publicly available high-dimensional biological microarray datasets for cancer diagnostics, in comparison to several conventional feature selection methods.

The remainder of this article is organized as follows. In the upcoming Section~\ref{sec:relatedwork} we briefly present related work on feature selection and feature transformation based dimensionality reduction approaches, as well as some recent information theoretic neural network training studies. We then describe the proposed MMINet approach on feature transformation learning neural networks with maximum mutual information criterion in Section~\ref{sec:mminet}. As part of our experimental studies in Section~\ref{sec:expstudies}, we initially illustrate the limitations of a simple feature selection approach with a toy example in Section~\ref{sec:toy}. In Section~\ref{sec:expdata} we describe both the synthetically generated and the diagnostic biological data sets that we used in our empirical assessments. Subsequently we describe our implementations and present our results in Sections~\ref{sec:implementation} and \ref{sec:results}. We conclude the article with a discussion of our methodology, results, current limitations and potential improvements.

\section{Related Work}
\label{sec:relatedwork}

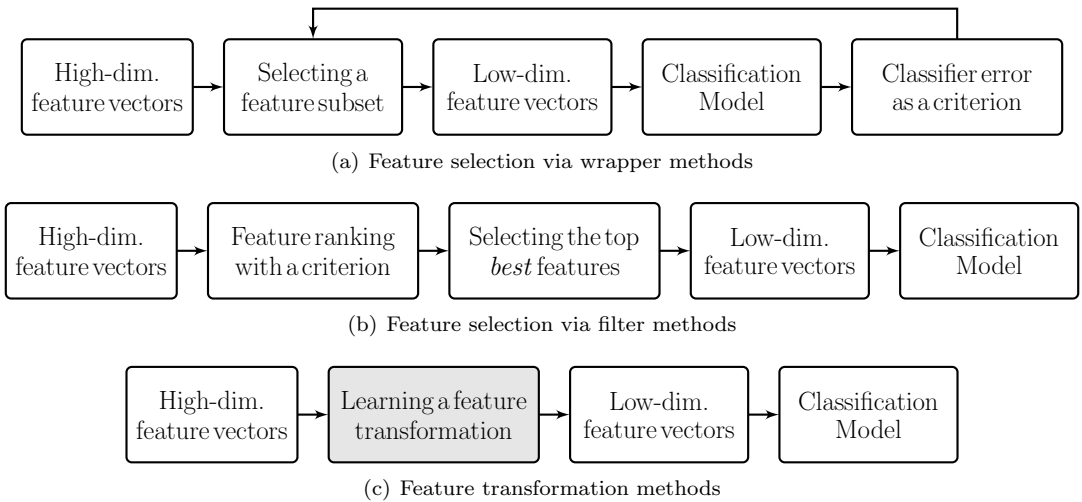
\begin{figure}
    \centering
    \subfigure[Feature selection via wrapper methods]
    {\begin{tikzpicture}[thick,scale=0.4, every node/.style={transform shape}]
    \tikzstyle{sblock} = [cloud, rectangle, rounded corners=2pt, draw, fill=white!10, text width=16em, minimum height=9em, text centered]
    \tikzstyle{block} = [cloud, rectangle, rounded corners=2pt, draw, fill=white!10, text width=19em, minimum height=9em, text centered]
    \node [sblock, inner sep=0pt] (HD) {\Huge High-dim.\\\vspace{0.2cm}feature vectors};
    \node [sblock, right = 1cm of HD] (SEL) {\Huge Selecting a\\\vspace{0.2cm}feature subset};
    \node [sblock, right = 1cm of SEL] (LD) {\Huge Low-dim.\\\vspace{0.3cm}feature vectors};
    \node [sblock, right = 1cm of LD] (CLS) {\Huge Classification\\\vspace{0.3cm}Model};
    \node [block, right = 1cm of CLS] (CLA) {\Huge Classifier error\\\vspace{0.3cm}as a criterion};
    \draw [draw, -latex'] (HD) -- (SEL);
    \draw [draw, -latex'] (SEL) -- (LD);
    \draw [draw, -latex'] (LD) -- (CLS);
    \draw [draw, -latex'] (CLS) -- (CLA);
    \draw [draw, -latex'] (CLA) |-  ([yshift=10mm] LD.north) -| (SEL);
    \end{tikzpicture}\label{fig:featwrapper}}
    
	\subfigure[Feature selection via filter methods]
	{\begin{tikzpicture}[thick,scale=0.4, every node/.style={transform shape}]
    \tikzstyle{sblock} = [cloud, rectangle, rounded corners=2pt, draw, fill=white!10, text width=16em, minimum height=9em, text centered]
    \tikzstyle{block} = [cloud, rectangle, rounded corners=2pt, draw, fill=white!10, text width=19em, minimum height=9em, text centered]
    \node [sblock, inner sep=0pt] (HD) {\Huge High-dim.\\\vspace{0.2cm}feature vectors};
    \node [block, right = 1cm of HD] (RNK) {\Huge Feature ranking\\\vspace{0.2cm}with a criterion};
    \node [block, right = 1cm of RNK] (SEL) {\Huge Selecting the top\\\vspace{0.2cm}\textit{best} features};
    \node [sblock, right = 1cm of SEL] (LD) {\Huge Low-dim.\\\vspace{0.3cm}feature vectors};
    \node [sblock, right = 1cm of LD] (CLS) {\Huge Classification\\\vspace{0.3cm}Model};
    \draw [draw, -latex'] (HD) -- (RNK);
    \draw [draw, -latex'] (RNK) -- (SEL);
    \draw [draw, -latex'] (SEL) -- (LD);
    \draw [draw, -latex'] (LD) -- (CLS);
    \end{tikzpicture}\label{fig:featfilter}}
    
	\subfigure[Feature transformation methods]
	{\begin{tikzpicture}[thick,scale=0.4, every node/.style={transform shape}]
    \tikzstyle{sblock} = [cloud, rectangle, rounded corners=2pt, draw, fill=white!10, text width=16em, minimum height=9em, text centered]
    \tikzstyle{block} = [cloud, rectangle, rounded corners=2pt, draw, fill=white!10, text width=19em, minimum height=9em, text centered]
    \node [sblock, inner sep=0pt] (HD) {\Huge High-dim.\\\vspace{0.2cm}feature vectors};
    \node [block, fill=gray!20, right = 1cm of HD] (TRS) {\Huge Learning a feature\\\vspace{0.2cm}transformation};
    \node [sblock, right = 1cm of TRS] (LD) {\Huge Low-dim.\\\vspace{0.3cm}feature vectors};
    \node [sblock, right = 1cm of LD] (CLS) {\Huge Classification\\\vspace{0.3cm}Model};
    \draw [draw, -latex'] (HD) -- (TRS);
    \draw [draw, -latex'] (TRS) -- (LD);
    \draw [draw, -latex'] (LD) -- (CLS);
    \end{tikzpicture}\label{fig:feattrans}}
    
    \caption{An illustration of common supervised dimensionality reduction approaches: (a) feature selection with wrapper methods which are particularly tailored for a classification model, (b) feature selection via filter methods which generally consider ranking and selection of features based on a pre-defined criterion independent of the classification model, (c) feature transformation approaches which aim to learn a mapping function based on an optimality criterion independent of the classification model.}
    \label{fig:selectiondiagrams}
\end{figure}

Supervised dimensionality reduction by feature selection refers to selecting the most class-informative feature subset from a high-dimensional feature set based on a defined optimality criterion to maximize class separability \cite{Guyon:2003}. A theoretically optimal dimensionality reduction procedure for a specified classifier is to iteratively adjust a pre-determined feature dimensionality reduction framework until the best cross validated decoding accuracy is achieved, which are known as the \textit{wrapper} methods (see Figure~\ref{fig:featwrapper}). One well-known example is the support vector machine (SVM) recursive feature elimination (RFE) approach \cite{Guyon:2002}. SVM-RFE is a wrapper feature selection method around an SVM classifier which uses backward elimination of features with the smallest model weights. Intuitively, as the dimensionality and amount of training data increases, wrapper methods become computationally cumbersome and time consuming for model learning. \textit{Filter} methods provide an alternative in the form of feature ranking and subset selection algorithms based on a pre-defined optimality criterion (see Figure~\ref{fig:featfilter}). In particular, feature selection based on information theoretic criteria, where salient statistical properties of features can be exploited by a probabilistic dependence measure, have shown significant promise in supervised dimensionality reduction \cite{Battiti:1994,Kwak:2002,Peng:2005}.

Feature selection methods offer the advantage of preserving original representations of the variables. This subsequently translates to sustaining better and easier model interpretability, and makes them preferable depending on the learning application domain \cite{Garrett:2003,Lazar:2012}. Nevertheless there exists significant evidence on feature ranking and selection algorithms leading to potentially sub-optimal solutions for class separability \cite{Erdogmus:2008,Torkkola:2008}. This argument can be simply illustrated by considering the case where two redundant features can become informative jointly (as will be shown in Section~\ref{sec:toy}). Accordingly, feature transformation based dimensionality reduction methods can provide a more robust and viable alternative \cite{Guyon:2003,Hinton:2006} (see Figure~\ref{fig:feattrans}), which are also demonstrated in the form of information theoretic linear projections or rotations \cite{Torkkola:2003,Hild:2006,Nenadic:2007,Zhang:2010,Faivishevsky:2012}. We motivate our study in the light of these work, where we aim to use standard gradient descent based artificial neural network training and inference pipelines to perform nonlinear maximum mutual information based feature transformations. We previously explored this idea for neurophysiological feature transformations in brain-computer interfaces \cite{Ozdenizci:2019}, which we re-address here in the context of neural networks.

Recently a different line of work focused on estimating mutual information of high dimensional continuous variables over neural networks, initially proposed as mutual information neural estimation (MINE) \cite{Belghazi:2018}. From an unsupervised representation learning perspective \cite{Hjelm:2018} extended MINE to learn powerful lower dimensional data representations that perform well on a variety of tasks, by maximizing the estimated mutual information between the input and output of a deep neural network encoder. More recently \cite{Wen:2020} proposed to estimate the gradient of mutual information rather than itself for similar representation learning setups, which was argued to provide a more stable estimate for unsupervised representation learning. Yet, these studies are particularly interested in learning unsupervised deep representations of continuous high-dimensional random variables from an information theoretic perspective, which are however being successfully translated into the convention of artificial neural networks.

Going further towards application domains, neural network based information theoretic metric estimators also demonstrated significant promise in various uses within diverse artificial intelligence settings. One of such use cases include medical dialogue systems for automatic diagnosis \cite{Xia:2020}, where mutual information estimation models are embedded within a policy learning framework to enhance the reward function and encourage the model to select the most discriminative symptoms to make a diagnosis. Another example extends disentangled representation learning models by an information theoretic formulation for image classification and retrieval problems in computer vision \cite{Sanchez:2019}. Potential contemporary use cases can further extend to mobile cloud computing applications \cite{Ciobanu:2019}, as well as end-to-end deep learning models for communication systems with efficient mutual information based encoding \cite{Fritschek:2019}.

\section{MMINet: Information Theoretic Dimensionality Reduction Neural Network}
\label{sec:mminet}

\subsection{Problem Statement}
\label{sec:formulation}

Let $\{(\bm{x}_i,c_i)\}_{i=1}^{n}$ denote the finite training data set where $\bm{x}_i\in\mathbb{R}^{d_x}$ is a sample of a continuous valued random variable $\mathit{X}$, and $c_i\in\{1,\ldots,L\}$ is a sample of a discrete valued random variable $\mathit{C}$, indicating the discrete class label for $\bm{x}_i$. From a dimensionality reduction perspective, the objective is to find a mapping network $\varphi^\star:\mathbb{R}^{d_x}\mapsto\mathbb{R}^{d_y}$ such that the high $d_x$-dimensional input feature space is mapped to a lower $d_y$-dimensional transformed feature space while maximizing the mutual information between the transformed data and corresponding class labels based on the observations, as expressed by Equation~\eqref{eq:objective}.
\begin{equation}
    \varphi^\star = \argmax_{\varphi\in\Omega} \{I(\mathit{Y},\mathit{C})\},
    \label{eq:objective}
\end{equation}
where the continuous random variable $\mathit{Y}$ has transformed data samples $\bm{y}_i=\varphi^\star(\bm{x}_i;\bm{\theta}^\star)$ in a $d_y$-dimensional feature space, $\bm{\theta}$ denotes the parameters of the mapping $\varphi$, and $\Omega$ denotes the function space for possible feature mappings $\varphi$.

In Bayesian optimal classification, upper and lower bounds on the probability of error in estimating a discrete valued random variable $\mathit{C}$ from an observational random variable $\mathit{Y}$ can be determined by information theoretic criteria (i.e., Fano's lower bound inequality \cite{Fano:1961} and Hellman-Raviv upper bound on Bayes error \cite{Hellman:1970}). Specifically, these bounds suggest that the lowest possible Bayes error of any given classifier can be achieved when the mutual information between the random variables $\mathit{Y}$ and $\mathit{C}$ is maximized (cf. \cite{Ozdenizci:2019,Torkkola:2003}).

\subsection{Learning with Maximum Mutual Information Criterion}
\label{sec:learning}

Mutual information between the continuous random variable $\mathit{Y}$ and the discrete random variable $\mathit{C}$ is defined as: $I(\mathit{Y},\mathit{C}) = H(\mathit{Y}) - H(\mathit{Y}\vert\mathit{C})$, which also can be expressed by Equation~\eqref{eq:mutualinfo}.
\begin{equation}
    \begin{split}
    I(\mathit{Y},\mathit{C}) = - \int_{\bm{y}} p(\bm{y})\log p(\bm{y})d\bm{y} + \int_{\bm{y}} \sum_{c} p(\bm{y},c)\log p(\bm{y} \vert c)d\bm{y}.
    \end{split}
    \label{eq:mutualinfo}
\end{equation}

\begin{figure}
    \centering
    \begin{tikzpicture}[thick,scale=0.4, every node/.style={transform shape}]
    \tikzstyle{empty} = [cloud, rectangle, text width=20em, text centered, minimum height=6em]
    \tikzstyle{sblock} = [cloud, rectangle, rounded corners=3pt, draw, fill=white!10, text width=20em, minimum height=9em, text centered]
    \tikzstyle{lblock} = [cloud, rectangle, rounded corners=3pt, draw, fill=white!10, text width=18em, minimum height=9em, text centered]
    \tikzstyle{gmghost} = [cloud, rectangle, rounded corners=3pt, draw, text width=15em, minimum height=9em, text centered]
    \tikzstyle{gmblock} = [cloud, rectangle, rounded corners=3pt, draw, fill=gray!20, text width=15em, minimum height=26em, text centered]
    \tikzstyle{loss} = [cloud,  rectangle, rounded corners=3pt, dashed, draw, fill=white!10, text width=16em, minimum height=9em, text centered]
    \tikzstyle{arrow} = [draw, -latex']
    \tikzstyle{eqtext} = [cloud, rectangle, text width=80em, text centered, minimum height=4em]

    \node [sblock] (HD) {\Huge High-dim.\\\vspace{0.3cm} feature vector: $\bm{x}_t$};
    \node [empty, above = 0cm of HD] (DUMMY) {};
    \node [sblock, above = 0cm of DUMMY] (TR) {\Huge Training Set\\\vspace{0.3cm}$\{(\bm{x}_i,c_i)\}_{i=1,i \ne t}^{n}$};
    \node [gmghost, right = 1.7cm of HD] (G1) {};
    \node [gmghost, right = 1.7cm of TR] (G2) {};
    \node [gmblock, right = 1.7cm of DUMMY] (TRS) {\Huge MMINet\\\vspace{0.4cm} $\bm{y}=\varphi(\bm{x};\bm{\theta})$};
    \node [sblock, right = 1.2cm of G1] (LD) {\Huge Low-dim.\\\vspace{0.4cm}feature vector: $\bm{y}_t$};
    \node [sblock, right = 1.2cm of G2] (PR) {\Huge Transformed Set\\\vspace{0.5cm}$\{(\bm{y}_i,c_i)\}_{i=1,i \ne t}^{n}$};
    \node [lblock, right = 1.2cm of LD] (MI) {\Huge Inst. Loss\\\vspace{0.4cm}$-\widehat{I}_t(\mathit{Y},\mathit{C})$};
    \node [loss, below = 1cm of LD] (Grad) {\Huge SMIG\\\vspace{0.4cm}$-\nabla_{\bm{\theta}}\widehat{I}_t(\mathit{Y},\mathit{C})$};
    
    \draw [arrow] (TR) -- (G2);
    \draw [arrow] (G2) -- (PR);
    \draw [arrow] (PR) -| (MI);
    \draw [arrow] (HD) -- (G1);
    \draw [arrow] (G1) -- (LD);
    \draw [arrow] (LD) -- (MI);
    \draw [arrow,dashed] (MI) |- (Grad);
    \draw [arrow,dashed] (Grad) -| (TRS);
    \end{tikzpicture}
    \caption{Stochastic training flow of MMINet which uses instantaneous training data samples $\bm{x}_t$ to calculate the instantaneous loss $-\widehat{I}_t(\mathit{Y},\mathit{C})$, and perform parameter updates based on its gradient (i.e., SMIG). Note that at every iteration $t$, the current transformed set samples based on the current $\bm{\theta}$ estimate are also needed to evaluate the instantaneous loss in Equation~\eqref{eq:instloss}.}
    \label{fig:itfldiagram}
\end{figure}
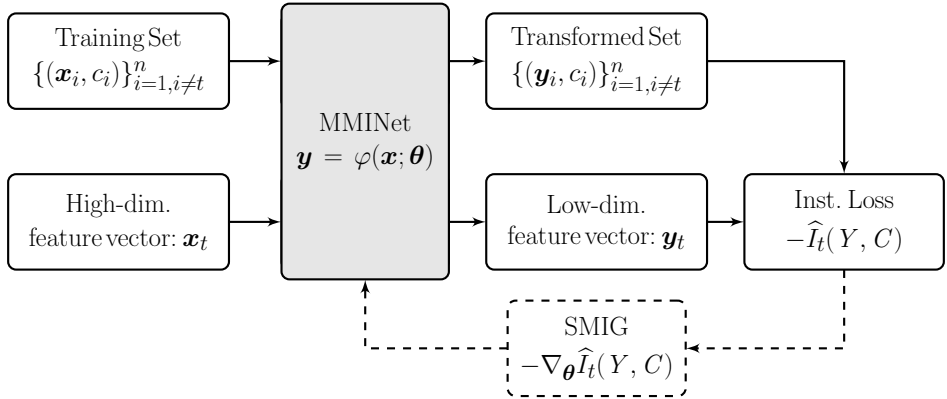

To solve the objective in Equation~\eqref{eq:objective}, exact estimation of the mutual information quantity is not necessary. Instead, we are only interested in adaptively estimating the optimal feature mapping network parameters $\bm{\theta}$ under maximum mutual information criterion. Motivated by similar work from information theory \cite{Erdogmus:2003,Chen:2008}, we approach the optimization problem stochastically. As illustrated in Figure~\ref{fig:itfldiagram}, the network parameters $\bm{\theta}$ will be iteratively updated based on the instantaneous estimate of the gradient of mutual information at each iteration $t$ (i.e., $\nabla_{\bm{\theta}}\widehat{I}_t(\mathit{Y},\mathit{C})$), which we define as the \textit{stochastic mutual information gradient (SMIG)}. 

During this network training procedure, in fact we approximate the true gradient of the mutual information $\nabla_{\bm{\theta}}I(Y,C)$ stochastically, and perform parameter updates based on the SMIG $\nabla_{\bm{\theta}}\widehat{I}_t(\mathit{Y},\mathit{C})$ evaluated with the instantaneous sample $\bm{y}_t$ and the values of $\bm{\theta}$ at iteration $t$. This stochastic estimate quantity can be obtained by dropping the expectation operation over $\mathit{Y}$ from the true gradient given in Equation~\eqref{eq:mutualinfograd}.
\begin{equation}
    \begin{split}
    \nabla_{\bm{\theta}}I(\mathit{Y},\mathit{C}) = \frac{\partial}{\partial\bm{\theta}} \Bigg[ - \int_{\bm{y}} p(\bm{y})\log p(\bm{y})d\bm{y} + \int_{\bm{y}} p(\bm{y}) \sum_{c} P(c \vert \bm{y})\log p(\bm{y} \vert c)d\bm{y} \Bigg].
    \end{split}
    \label{eq:mutualinfograd}
\end{equation}

Subsequently, the expression for SMIG at iteration $t$ can be denoted by Equation~\eqref{eq:smig}.
\begin{equation}
    \begin{split}
    \nabla_{\bm{\theta}}\widehat{I}_t(\mathit{Y},\mathit{C}) = \frac{\partial}{\partial\bm{\theta}} \Bigg[- \log \widehat{p}(\bm{y}_t) +  \sum_{c} \widehat{P}(c \vert \bm{y}_t) \log \widehat{p}(\bm{y}_t \vert c)\Bigg].
    \end{split}
    \label{eq:smig}
\end{equation}

In the neural network training process, consistently with Figure~\ref{fig:itfldiagram}, we simply use $-\widehat{I}_t(\mathit{Y},\mathit{C})$ as the instantaneous loss to be backpropagated over the network for parameter updates at iteration $t$. Applying the Bayes' Theorem, the instantaneous loss estimate from Equation~\eqref{eq:smig} can be expressed via Equation~\eqref{eq:instloss}.
\begin{equation}
    \begin{split}
    -\widehat{I}_t(\mathit{Y},\mathit{C}) = + \log \left( \sum_{c} \widehat{P}(c)\widehat{p}(\bm{y}_t \vert c) \right) - \sum_{c} \left(\frac{\widehat{P}(c)\widehat{p}(\bm{y}_t \vert c)}{\sum_{c} \widehat{P}(c)\widehat{p}(\bm{y}_t \vert c)} \right) \log \widehat{p}(\bm{y}_t \vert c),
    \end{split}
    \label{eq:instloss}
\end{equation}
where the class priors $\widehat{P}(c)$ will be empirically determined over the training data samples, and $\widehat{p}(\bm{y}_t \vert c)$ at each iteration $t$ will be approximated via non-parametric kernel density estimations \cite{Principe:2000} on class conditional distributions of the transformed data samples expressed as in Equation~\eqref{eq:kde}.
\begin{equation}
    \begin{split}
    \widehat{p}(\bm{y}_t \vert c) = \frac{1}{n_c}\sum_{j=1}^{n_c} \textbf{K}_{\textbf{H}}(\bm{y}_t-\bm{y}_j),
    \end{split}
    \label{eq:kde}
\end{equation}
where index $j$ iterates over the training samples of the conditioned class $c$ and $n_c$ denotes the number of samples in that class. Since a continuously differentiable kernel choice is necessary for proper evaluation of the gradients, we use Gaussian kernels as denoted in Equation~\eqref{eq:gaussiankernel}.
\begin{equation}
    \begin{split}
    \textbf{K}_{\textbf{H}}(\bm{y}_t-\bm{y}_j)=\frac{1}{(2\pi)^{d_y/2}|\textbf{H}|^{1/2}}e^{\frac{1}{2}(\bm{y}_t-\bm{y}_j)^{\text{T}}\textbf{H}^{-1}(\bm{y}_t-\bm{y}_j)},
    \end{split}
    \label{eq:gaussiankernel}
\end{equation}
with the kernel bandwidth matrix $\textbf{H}$ determined using Silverman's rule of thumb \cite{Silverman:1986}. Finally, note that the SMIG in Equation~\eqref{eq:smig} is a biased estimator of the true gradient of mutual information in Equation~\eqref{eq:mutualinfograd}, since it is based on kernel density estimators with finite samples which are biased estimators \cite{Parzen:1962}. An increase in the training data sample size per class can yield better class conditional kernel density estimates \cite{Hwang:1994} that can be exploited during the neural network optimization process.

\section{Experimental Studies}
\label{sec:expstudies}

\subsection{An Illustrative Example}
\label{sec:toy}

\begin{figure}
	\centering
	\includegraphics[clip, trim=0.5cm 0cm 0.3cm 0.1cm, width=0.26\textwidth]{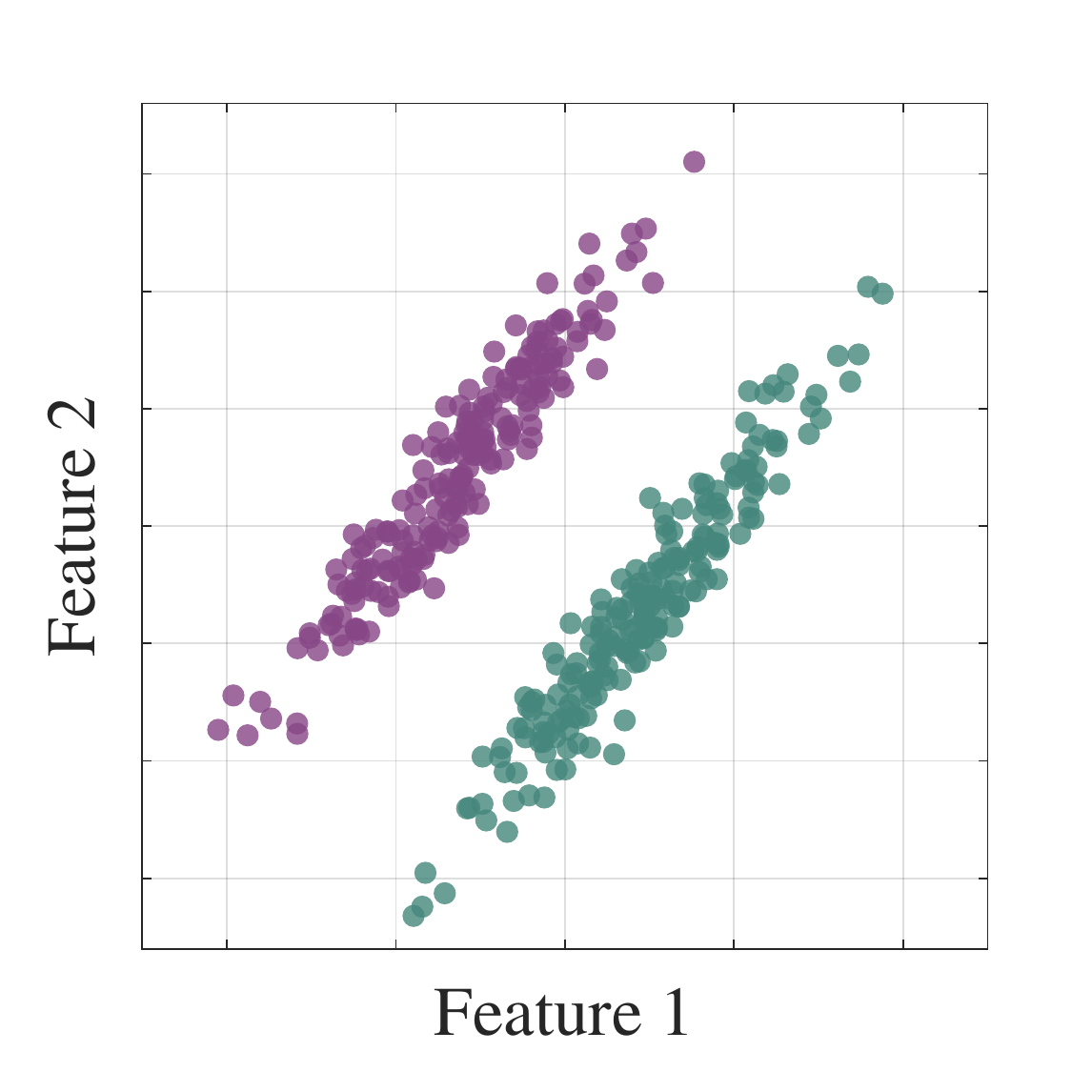}\hspace{-0.3cm}
	\includegraphics[clip, trim=0.5cm 0cm 0.3cm 0.1cm,width=0.26\textwidth]{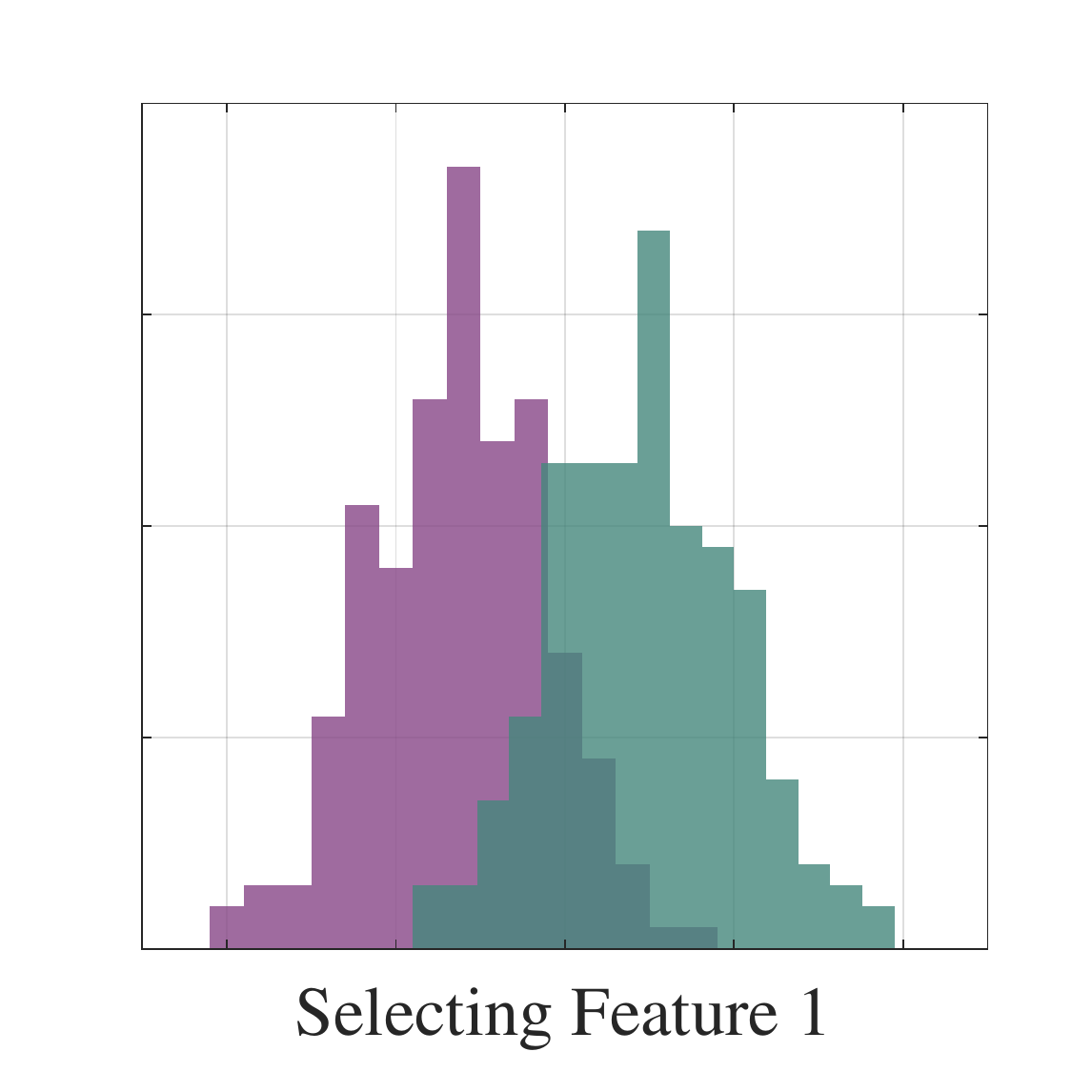}\hspace{-0.3cm}
	\includegraphics[clip, trim=0.5cm 0cm 0.3cm 0.1cm,width=0.26\textwidth]{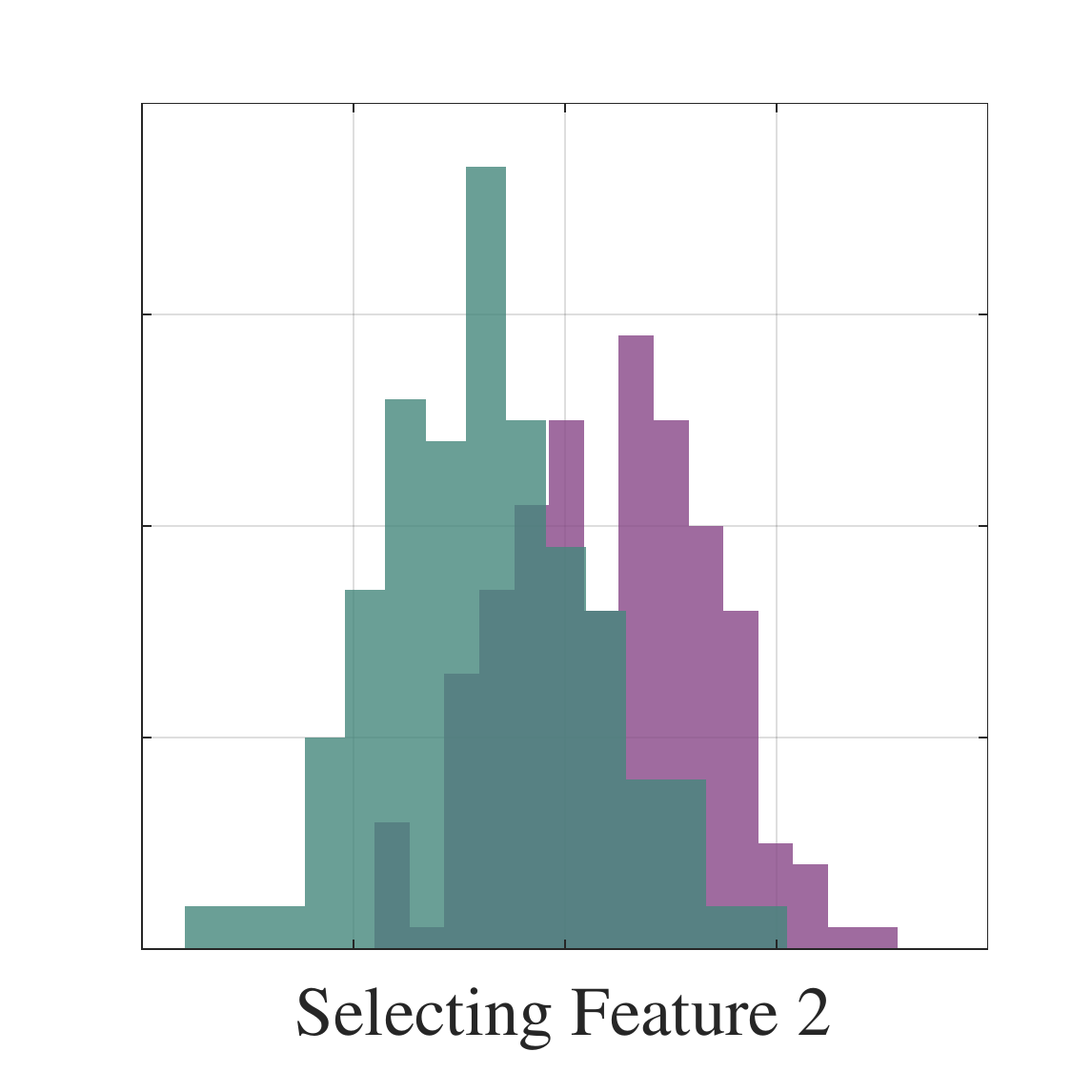}\hspace{-0.3cm}
	\includegraphics[clip, trim=0.5cm 0cm 0.3cm 0.1cm,width=0.26\textwidth]{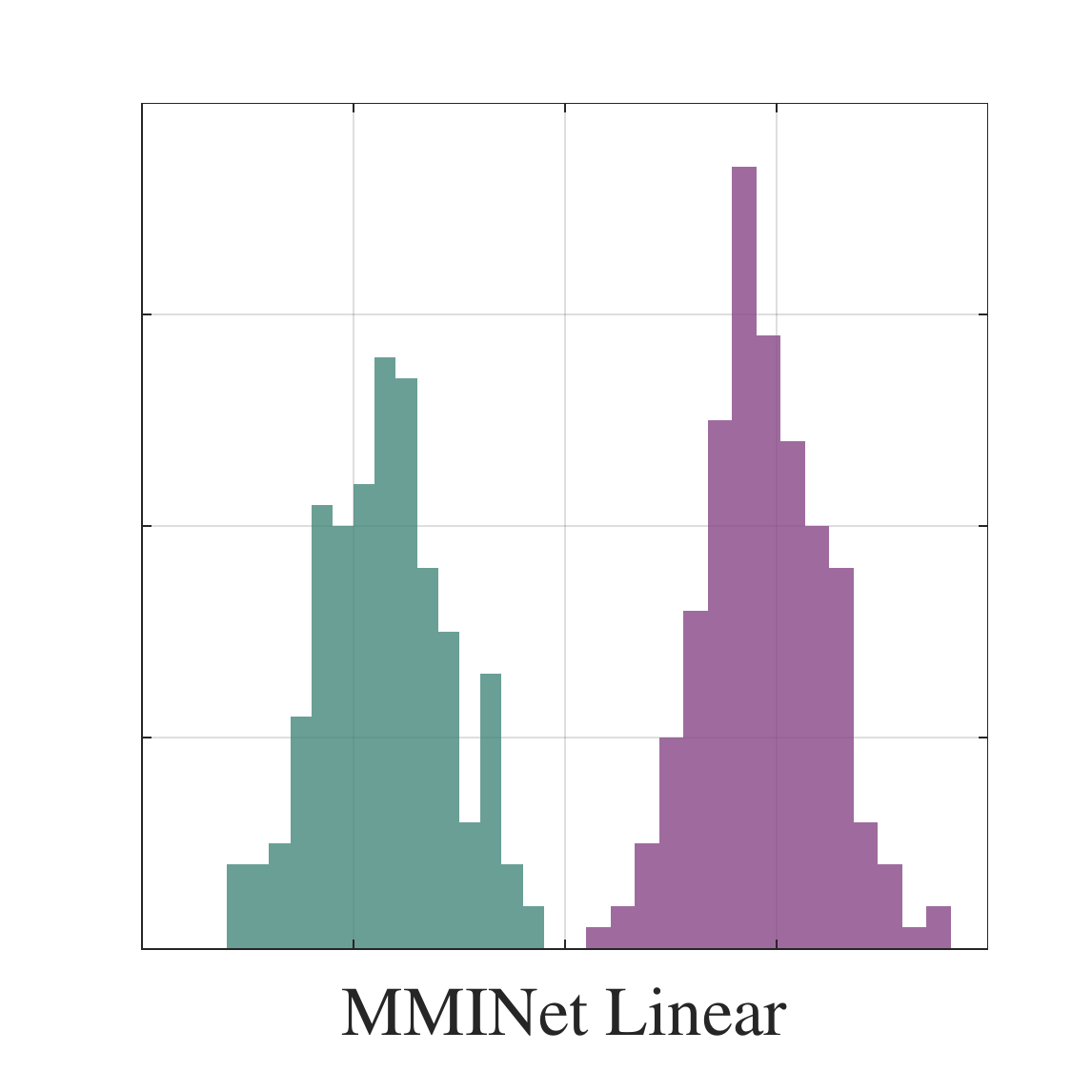}\hspace{-0.3cm}
	\caption{An illustration of how feature selection can confound class separability, using a two-dimensional data distribution from two color-coded classes as demonstrated at the top left figure. For the two classes, there is significant overlap in distributions when an individual feature is selected. Bottom right figure shows single linear layer MMINet projections onto a one dimension.}
	\label{fig:toy_results}
\end{figure}

We first demonstrate a simple example on how feature selection can lead to confounding results regarding class separability as we highlighted in Section~\ref{sec:intro}. We will illustrate a two-class classification problem with two-dimensional data distributions such that there is significant overlap in distributions when an individual feature is selected (see Figure~\ref{fig:toy_results}). While class distributions are easily separable when both features are considered together, a feature selection between the two dimensions will lead to significant information loss. 

We subsequently show the projection results using a simple MMINet architecture with a single linear (dense) layer $\varphi(\bm{x};\bm{\theta})=\bm{W}\bm{x}$, where $\bm{W}$ is a one by two projection array. We observe that maximum mutual information criterion based linear feature transformation ensures minimum probability of error based on the available training data samples. This example illustrates one setting on how feature selection can lead to sub-optimal solutions for class separability.

\subsection{Experimental Data}
\label{sec:expdata}

We evaluate our information theoretic dimensionality reduction approach on two different types of datasets. Firstly we perform feasibility assessments on a synthetically generated dataset, and later conduct experiments using three diagnostic biological microarray datasets.

\subsubsection{Synthetically Generated Data}
\label{sec:synthetic}

Preliminary evaluations of our approach are performed using an artificially generated dataset with regards to a well-known basis for comparison of learning algorithms \cite{Thrun:1991}. We use the Monk3 Dataset, from the MONK's problems \cite{Thrun:1991}, which handles a binary classification task where 432 data samples are described by $d_x=6$ features $(x_1,\ldots,x_6)$. For each data sample, binary class labels are obtained by the following logical operation: $(x_5=3 \wedge x_4=1) \lor (x_5\ne 4 \wedge x_2\ne 3)$. From the 432 data samples, 5\% have noisy labels. Overall, the problem implies that there are only three relevant features $(x_2,x_4,x_5)$ to infer the class label and the remaining three features are redundant.

\subsubsection{High-Dimensional Diagnostic Biological Data}
\label{sec:biodata}

We perform further empirical assessments using high-dimensional biological microarray data from the following three datasets: (1) Breast Cancer Wisconsin Diagnostic Dataset \cite{Dua:2019} consisting of 569 samples of 30 dimensional features extracted from digitized images of a fine needle aspirate of a breast mass, describing cell characteristics where the cell is either classified as malignant or benign, (2) Glioma Dataset \cite{Nutt:2003} containing 50 samples of four class data (i.e., cancer/non-cancer glioblastomas or oligodendrogliomas) defined by high-dimensional microarray gene expression data of 4434 features, (3) Lung Carcinoma Dataset \cite{Bhattacharjee:2001} containing 203 samples in five classes adenocarcinomas, squamous cell lung carcinomas, pulmonary carcinoids, small-cell lung carcinomas and normal lung, defined by 3312 mRNA gene expression variables.

\subsection{Implementations}
\label{sec:implementation}

We evaluate MMINet feature transformation method in comparison to four supervised feature selection methods: (1) feature selection based on \textit{fisher score} as a similarity based approach \cite{Duda:2012}, (2) minimum redundancy maximum relevance (\textit{mRMR}) feature selection \cite{Peng:2005} as an information theoretic feature ranking and selection criterion, (3) \textit{$l_1$-SVM} as a sparse regularization based method that utilizes an $l_1$-norm regularizer on a linear support vector machine (SVM) classifier \cite{Zhu:2004}, (4) \textit{SVM-RFE} as a wrapper feature selection method around an SVM classifier with recursive feature elimination (RFE) where features with lowest SVM weights are recursively eliminated backwards \cite{Guyon:2002}. 

For our MMINet implementations\footnote{Codes are available at: https://github.com/oozdenizci/MMIDimReduction} we used the Chainer deep learning framework \cite{Tokui:2015}. Stochastic model training was performed by considering one instantaneous sample at a time (i.e., one sample per batch) for one complete pass across the whole training dataset (i.e., one epoch), and we employed momentum stochastic gradient descent \cite{Qian:1999}. The neural network architecture was arbitrarily defined as a two hidden layer network with ELU activations following the hidden layer outputs. Dimensionalities of the dense layers were chosen to be $d_x$ to $d_x/2$, $d_x/2$ to $d_x/4$, and $d_x/4$ to $d_y$. All features were standardized by removing the mean and scaling to unit variance. After dimensionality reduction (i.e., feature selection or MMINet transformation), for classification purposes we used linear SVM classifiers and reported averaged 5-fold cross-validation accuracies in all experiments.

\subsection{Results}
\label{sec:results}

Results with the synthetically generated Monk3 Dataset \cite{Thrun:1991} demonstrated higher accuracies with MMINet in several experiments. We performed dimensionality reduction (from $d_x=6$) and 5-fold cross-validated classification of the 432 data samples for output dimensionalities of $d_y\in\{1,2,3\}$. For $d_y=1$, MMINet yields the highest average accuracy of $80.81\%$, with regards to $63.87\%$ with mRMR and $72.21\%$ with the other methods. Considering that the dimensionality reduction problem handles six input features, we observed that several of the feature selection methods identified the same feature as the most informative to construct $d_y$, which consistently resulted in this $72.21\%$ decoding accuracy. Similar behavior is observed for $d_y=2$, where MMINet yields a $77.05\%$ accuracy, whereas all feature selection methods selected the same two features and yield an average accuracy of $75.92\%$. Finally for $d_y=3$, MMINet yields $76.63\%$, $l_1$-SVM and SVM-RFE yields $76.87\%$, and the other two methods yield an average accuracy of $76.93\%$. This indicates that for selection of three features, in almost all cross-validation folds feature selection methods choose the truly relevant features, while MMINet transformations also yield comparable results. The overall upper accuracy range is further dependent on our choice to use a linear classifier for this problem. We did not increase the output feature dimensionality higher than three due to the nature of the constructed artificial dataset.

\renewcommand{\arraystretch}{1.1}
\begin{table}
  \caption{Dataset descriptives and averaged 5-fold cross-validation classification accuracies (\%) after dimensionality reduction by feature selection or MMINet feature transformation, for specific $d_y$.}\vspace{0.2cm}
  \label{tab:bioresults}
  \centering
  \begin{tabular}{c c c c}
    \toprule
    \textbf{} & \textbf{Breast Cancer} \cite{Dua:2019} & \textbf{Glioma} \cite{Nutt:2003} & \textbf{Lung Carcinoma} \cite{Bhattacharjee:2001} \\
    \midrule
    number of classes & $2$ & $4$ & $5$  \\
    number of data samples & $569$ & $50$ & $203$ \\
    $d_x \rightarrow d_y$ & $30 \rightarrow 2$ & $4434 \rightarrow 4$ & $3312 \rightarrow 5$ \\
    \midrule
    Fisher Score & $93.50$ & $50.00$ & $76.85$ \\
    mRMR & $91.56$ & $42.00$ & $78.33$ \\
    $l_1$-SVM & $92.09$ & $34.00$ & $84.28$ \\
    SVM-RFE & $93.14$ & $56.00$ & $89.65$ \\
    \textbf{MMINet} & $\mathbf{94.73}$ & $\mathbf{64.00}$ & $\mathbf{92.61}$ \\
    \bottomrule
  \end{tabular}
\end{table}

Regarding our experiments with high-dimensional diagnostic biological data, Table~\ref{tab:bioresults} presents averaged 5-fold cross-validation accuracies for the cases where output feature dimensionality $d_y$ is chosen equal to the number of classes for consistency across methods. MMINet yields accuracies of $94.73\%$ for binary classification with the Breast Cancer Wisconsin Diagnostic Dataset \cite{Dua:2019}, $64.00\%$ for 4-class classification with the Glioma Dataset \cite{Nutt:2003}, and $92.61\%$ for 5-class classification with the Lung Carcinoma Dataset \cite{Bhattacharjee:2001}, all relatively higher than the compared feature selection methods. We observe that our feature transformation approach provides a performance upper bound to several feature selection methods in classification, based on the same classifier modality. We argue this to be due to feature selection algorithms being more restricted and simply resemble to sparse linear projection solutions when MMINet would be constrained to have a single dense layer. 

\begin{figure}
	\centering
	\subfigure[Breast Cancer \cite{Dua:2019}]{\includegraphics[width=0.33\textwidth]{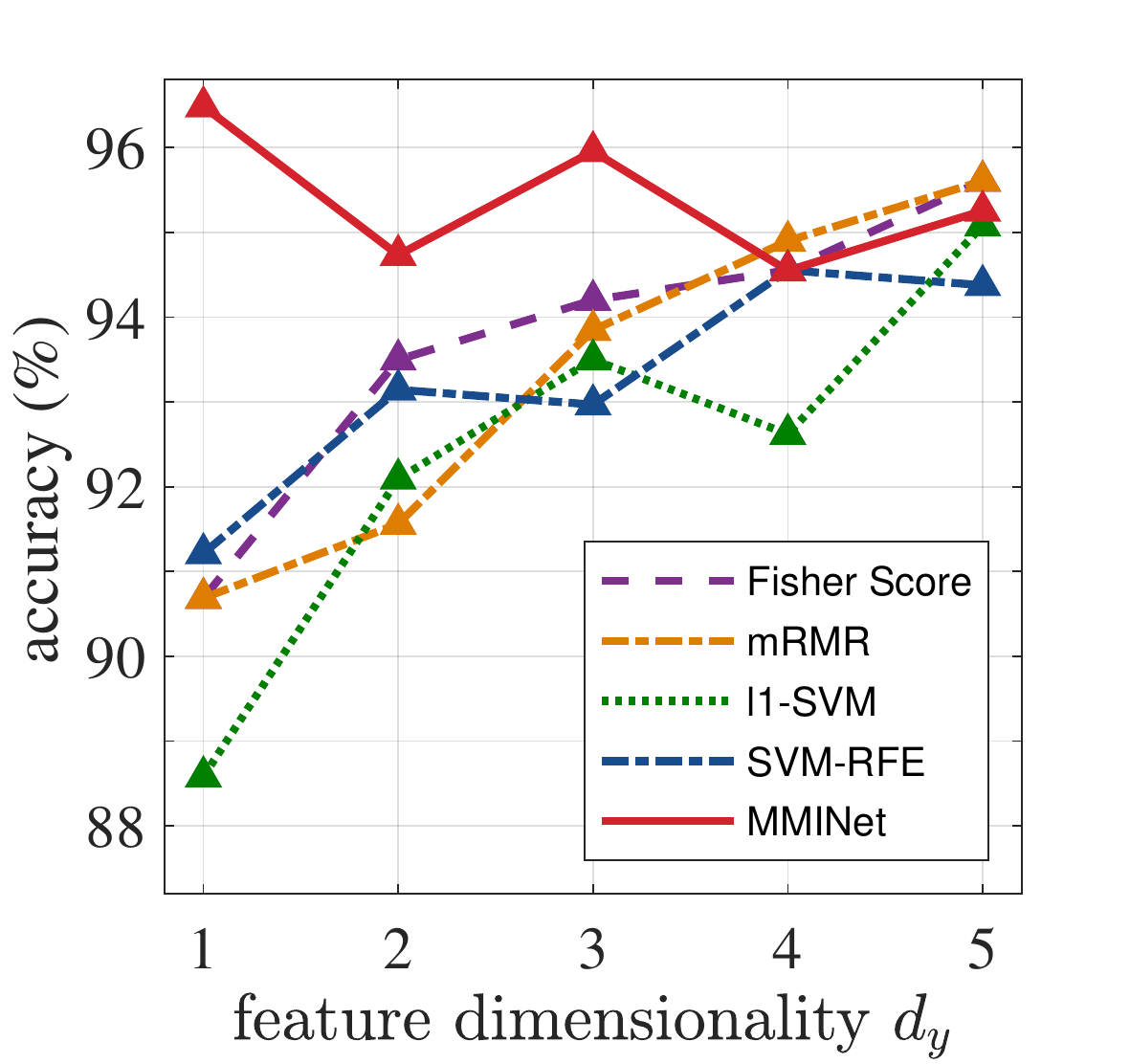}}\hspace{-0.1cm}
	\subfigure[Glioma \cite{Nutt:2003}]{\includegraphics[width=0.33\textwidth]{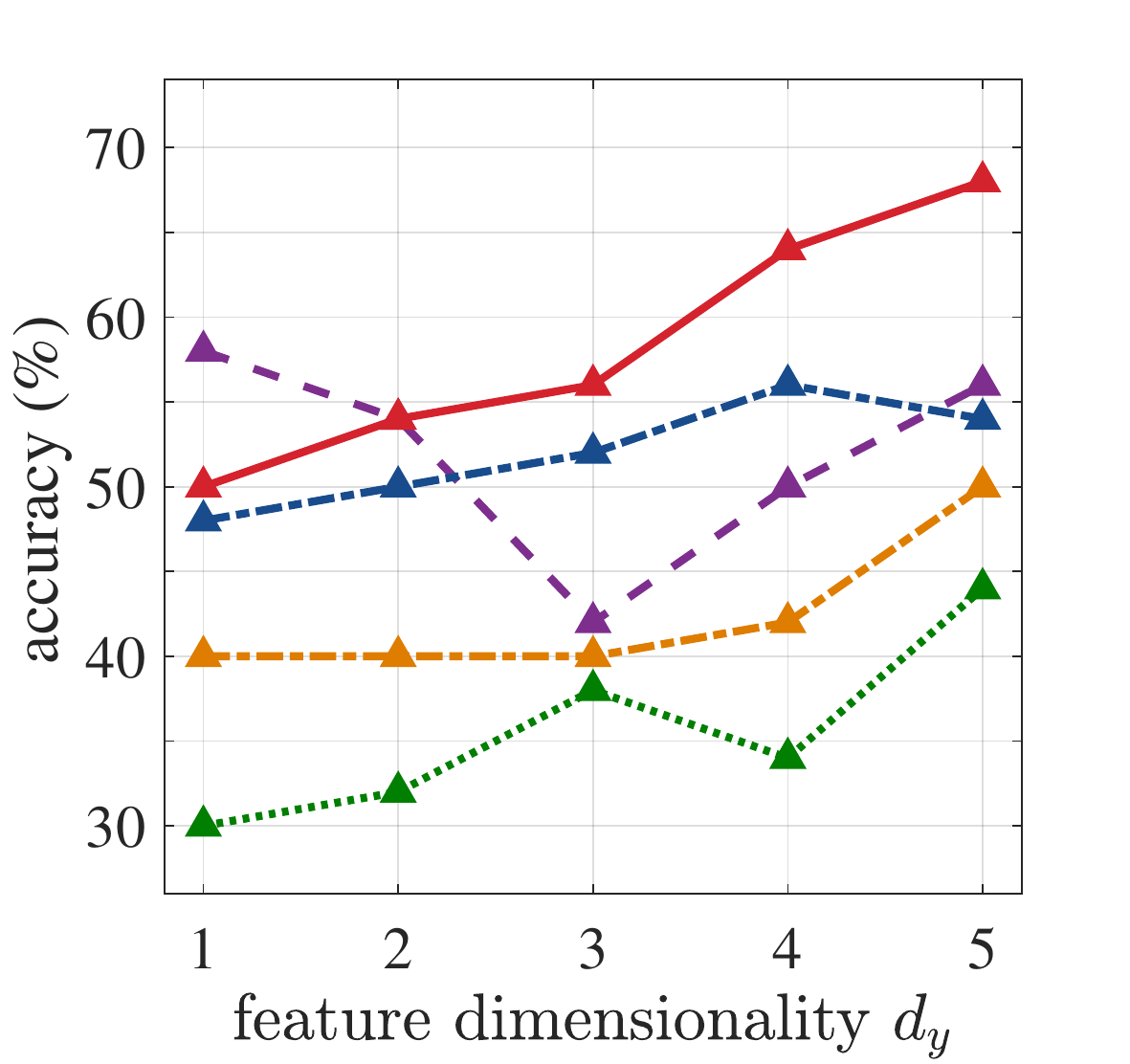}}\hspace{-0.1cm}
	\subfigure[Lung Carcinoma \cite{Bhattacharjee:2001}]{\includegraphics[width=0.33\textwidth]{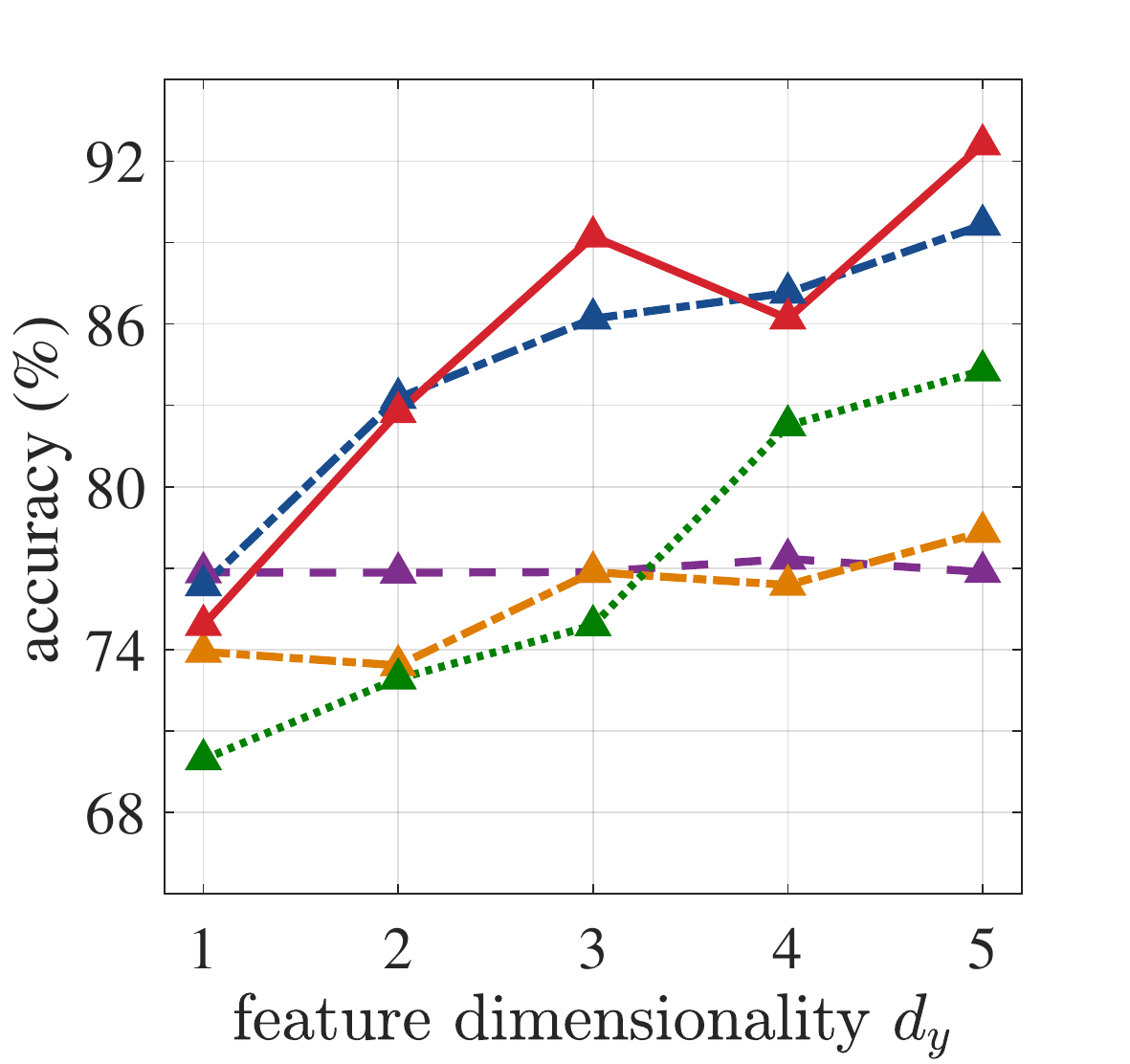}}\hspace{-0.1cm}
	\caption{Averaged 5-fold cross-validation classification accuracies (\%) for varying output feature dimensionalities ($d_y$) with all five dimensionality reduction methods and the three datasets. Line color and styles are consistent across all three plots as indicated in the legend for (a).}
	\label{fig:result_plots}
\end{figure}

Figure~\ref{fig:result_plots} demonstrates an extension of the results in Table~\ref{tab:bioresults}, where we vary the output feature dimensionality $d_y\in\{1,2,3,4,5\}$ for all datasets. We observe in almost all cases that MMINet continues to provide a better performance than the other methods. Mainly SVM-RFE, a wrapper method, is competitive with MMINet as anticipated due to the classifier-oriented nature of the algorithm. Note that we did not arbitrarily increase the number of dimensions too high for MMINet, since the method relies on $d_y$-dimensional kernel density estimators at the output feature space and higher dimensional density estimates are known to be unstable \cite{Silverman:1986}.

\section{Discussion}

We present a supervised dimensionality reduction network training procedure based on the stochastic estimate of the mutual information gradient. Based on the construction of the objective function, at the network output feature space the transformed features and their associated class labels carry maximum mutual information. Complete process is formulated non-parametrically based on kernel density estimates which approximate class-conditional densities in the projected feature space. We demonstrate our approach empirically using pilot experimental biological data, where feature selection algorithms are widely popular approaches for dimensionality reduction. We interpret our approach to be a more general solution than maximum mutual information based feature selection algorithms. Such selection algorithms resemble to sparse linear projection solutions when MMINet is constrained to have a single dense layer.

It is well known that the ultimate objective in Equation~\eqref{eq:objective} is hard to estimate due to entangling multiple continuous and discrete random variables where continuous random variables can have infinitely large positive or negative entropy values, whereas the entropy of a discrete random variable is always positive \cite{Ross:2014,Gao:2017}. Due to the fact that there is not a global solution to optimize this objective, it is important to note that the stopping criteria is an important factor in our model training. For our current implementations we did not optimize this aspect by using a validation set based stopping criterion, which could further improve the robustness of the approach.

We stress the importance of the distinction between our study and conventional discriminative neural network training protocols. Such discriminative networks are trained end-to-end using raw data to minimize negative log-likelihood as a measure of classification error minimization based on a training data set. On the other hand, our approach is a general supervised feature dimensionality reduction and lower-dimensional feature space learning method which relies on maximum mutual information criterion. Therefore we did not perform comparisons of the dimensionality reduction methods to discriminative neural networks such that a comparable basis is maintained.

Going beyond multilayer perceptrons, stochastic training of the MMINet framework can also embed any deep neural network architecture for lower dimensional representation learning. It is important to note that in contrast to feature selection methods, which preserve the original representations of feature variables, our transformation based approach will deeply modify features onto a new feature space. In combination with the theoretical advancements on gradient-based methods of neural network interpretability (e.g., layer-wise relevance propagation \cite{Bach:2015,Montavon:2018}), obtained synergies across features as highlighted by high-dimensional feature \textit{relevances} can yield significant insights based on the application domain. Such feature-synergy based ideas were particularly found interesting for feature learning in brain interfacing as we studied earlier \cite{Ozdenizci:2019,Ozdenizci:2020}, as well as gene expression data analysis \cite{Jacob:2009} in consistency with their biological interpretations.

\section*{Acknowledgments}
Our work is supported by NSF (IIS-1149570, CNS-1544895, IIS-1715858), DHHS (90RE5017-02-01), and NIH (R01DC009834).

\bibliographystyle{model5-names}

\begin{thebibliography}{48}
\expandafter\ifx\csname natexlab\endcsname\relax\def\natexlab#1{#1}\fi
\providecommand{\url}[1]{\texttt{#1}}
\providecommand{\href}[2]{#2}
\providecommand{\path}[1]{#1}
\providecommand{\DOIprefix}{doi:}
\providecommand{\ArXivprefix}{arXiv:}
\providecommand{\URLprefix}{URL: }
\providecommand{\Pubmedprefix}{pmid:}
\providecommand{\doi}[1]{\href{http://dx.doi.org/#1}{\path{#1}}}
\providecommand{\Pubmed}[1]{\href{pmid:#1}{\path{#1}}}
\providecommand{\bibinfo}[2]{#2}
\ifx\xfnm\relax \def\xfnm[#1]{\unskip,\space#1}\fi
\bibitem[{Bach et~al.(2015)Bach, Binder, Montavon, Klauschen, M{\"u}ller \&
  Samek}]{Bach:2015}
\bibinfo{author}{Bach, S.}, \bibinfo{author}{Binder, A.},
  \bibinfo{author}{Montavon, G.}, \bibinfo{author}{Klauschen, F.},
  \bibinfo{author}{M{\"u}ller, K.-R.}, \& \bibinfo{author}{Samek, W.}
  (\bibinfo{year}{2015}).
\newblock \bibinfo{title}{On pixel-wise explanations for non-linear classifier
  decisions by layer-wise relevance propagation}.
\newblock {\it \bibinfo{journal}{PloS one}\/},  {\it \bibinfo{volume}{10}\/},
  \bibinfo{pages}{e0130140}.
\bibitem[{Battiti(1994)}]{Battiti:1994}
\bibinfo{author}{Battiti, R.} (\bibinfo{year}{1994}).
\newblock \bibinfo{title}{Using mutual information for selecting features in
  supervised neural net learning}.
\newblock {\it \bibinfo{journal}{IEEE Transactions on Neural Networks}\/},
  {\it \bibinfo{volume}{5}\/}, \bibinfo{pages}{537--550}.
\bibitem[{Belghazi et~al.(2018)Belghazi, Baratin, Rajeshwar, Ozair, Bengio,
  Courville \& Hjelm}]{Belghazi:2018}
\bibinfo{author}{Belghazi, M.~I.}, \bibinfo{author}{Baratin, A.},
  \bibinfo{author}{Rajeshwar, S.}, \bibinfo{author}{Ozair, S.},
  \bibinfo{author}{Bengio, Y.}, \bibinfo{author}{Courville, A.}, \&
  \bibinfo{author}{Hjelm, D.} (\bibinfo{year}{2018}).
\newblock \bibinfo{title}{Mutual information neural estimation}.
\newblock In {\it \bibinfo{booktitle}{International Conference on Machine
  Learning}\/} (pp. \bibinfo{pages}{531--540}).
\bibitem[{Bhattacharjee et~al.(2001)}]{Bhattacharjee:2001}
\bibinfo{author}{Bhattacharjee, A.} et~al. (\bibinfo{year}{2001}).
\newblock \bibinfo{title}{Classification of human lung carcinoma by
  m\uppercase{RNA} expression profiling reveals distinct adenocarcinoma
  subclasses}.
\newblock {\it \bibinfo{journal}{PNAS}\/},  {\it \bibinfo{volume}{98}\/},
  \bibinfo{pages}{13790--13795}.
\bibitem[{Chen et~al.(2008)Chen, Hu, Li \& Sun}]{Chen:2008}
\bibinfo{author}{Chen, B.}, \bibinfo{author}{Hu, J.}, \bibinfo{author}{Li, H.},
  \& \bibinfo{author}{Sun, Z.} (\bibinfo{year}{2008}).
\newblock \bibinfo{title}{Adaptive filtering under maximum mutual information
  criterion}.
\newblock {\it \bibinfo{journal}{Neurocomputing}\/},  {\it
  \bibinfo{volume}{71}\/}, \bibinfo{pages}{3680--3684}.
\bibitem[{Ciobanu et~al.(2019)Ciobanu, Dobre, B{\u{a}}l{\u{a}}nescu \&
  Suciu}]{Ciobanu:2019}
\bibinfo{author}{Ciobanu, R.-I.}, \bibinfo{author}{Dobre, C.},
  \bibinfo{author}{B{\u{a}}l{\u{a}}nescu, M.}, \& \bibinfo{author}{Suciu, G.}
  (\bibinfo{year}{2019}).
\newblock \bibinfo{title}{Data and task offloading in collaborative mobile
  fog-based networks}.
\newblock {\it \bibinfo{journal}{IEEE Access}\/},  {\it \bibinfo{volume}{7}\/},
  \bibinfo{pages}{104405--104422}.
\bibitem[{Dua \& Graff(2019)}]{Dua:2019}
\bibinfo{author}{Dua, D.}, \& \bibinfo{author}{Graff, C.}
  (\bibinfo{year}{2019}).
\newblock \bibinfo{title}{{UCI} machine learning repository}.
\newblock \URLprefix \url{http://archive.ics.uci.edu/ml}.
\bibitem[{Duda et~al.(2012)Duda, Hart \& Stork}]{Duda:2012}
\bibinfo{author}{Duda, R.~O.}, \bibinfo{author}{Hart, P.~E.}, \&
  \bibinfo{author}{Stork, D.~G.} (\bibinfo{year}{2012}).
\newblock {\it \bibinfo{title}{Pattern Classification}\/}.
\newblock \bibinfo{publisher}{John Wiley \& Sons}.
\bibitem[{Erdogmus et~al.(2003)Erdogmus, Hild \& Principe}]{Erdogmus:2003}
\bibinfo{author}{Erdogmus, D.}, \bibinfo{author}{Hild, K.~E.}, \&
  \bibinfo{author}{Principe, J.~C.} (\bibinfo{year}{2003}).
\newblock \bibinfo{title}{Online entropy manipulation: Stochastic information
  gradient}.
\newblock {\it \bibinfo{journal}{IEEE Signal Processing Letters}\/},  {\it
  \bibinfo{volume}{10}\/}, \bibinfo{pages}{242--245}.
\bibitem[{Erdogmus et~al.(2008)Erdogmus, Ozertem \& Lan}]{Erdogmus:2008}
\bibinfo{author}{Erdogmus, D.}, \bibinfo{author}{Ozertem, U.}, \&
  \bibinfo{author}{Lan, T.} (\bibinfo{year}{2008}).
\newblock \bibinfo{title}{Information theoretic feature selection and
  projection}.
\newblock In {\it \bibinfo{booktitle}{Speech, Audio, Image and Biomedical
  Signal Processing using Neural Networks}\/} (pp. \bibinfo{pages}{1--22}).
\bibitem[{Faivishevsky \& Goldberger(2012)}]{Faivishevsky:2012}
\bibinfo{author}{Faivishevsky, L.}, \& \bibinfo{author}{Goldberger, J.}
  (\bibinfo{year}{2012}).
\newblock \bibinfo{title}{Dimensionality reduction based on non-parametric
  mutual information}.
\newblock {\it \bibinfo{journal}{Neurocomputing}\/},  {\it
  \bibinfo{volume}{80}\/}, \bibinfo{pages}{31--37}.
\bibitem[{Fano(1961)}]{Fano:1961}
\bibinfo{author}{Fano, R.~M.} (\bibinfo{year}{1961}).
\newblock {\it \bibinfo{title}{Transmission of information: A statistical
  theory of communications}\/}.
\bibitem[{Fritschek et~al.(2019)Fritschek, Schaefer \& Wunder}]{Fritschek:2019}
\bibinfo{author}{Fritschek, R.}, \bibinfo{author}{Schaefer, R.~F.}, \&
  \bibinfo{author}{Wunder, G.} (\bibinfo{year}{2019}).
\newblock \bibinfo{title}{Deep learning for channel coding via neural mutual
  information estimation}.
\newblock In {\it \bibinfo{booktitle}{IEEE 20th International Workshop on
  Signal Processing Advances in Wireless Communications}\/} (pp.
  \bibinfo{pages}{1--5}).
\bibitem[{Gao et~al.(2017)Gao, Kannan, Oh \& Viswanath}]{Gao:2017}
\bibinfo{author}{Gao, W.}, \bibinfo{author}{Kannan, S.}, \bibinfo{author}{Oh,
  S.}, \& \bibinfo{author}{Viswanath, P.} (\bibinfo{year}{2017}).
\newblock \bibinfo{title}{Estimating mutual information for discrete-continuous
  mixtures}.
\newblock In {\it \bibinfo{booktitle}{Advances in Neural Information Processing
  Systems}\/} (pp. \bibinfo{pages}{5986--5997}).
\bibitem[{Garrett et~al.(2003)Garrett, Peterson, Anderson \&
  Thaut}]{Garrett:2003}
\bibinfo{author}{Garrett, D.}, \bibinfo{author}{Peterson, D.~A.},
  \bibinfo{author}{Anderson, C.~W.}, \& \bibinfo{author}{Thaut, M.~H.}
  (\bibinfo{year}{2003}).
\newblock \bibinfo{title}{Comparison of linear, nonlinear, and feature
  selection methods for \uppercase{EEG} signal classification}.
\newblock {\it \bibinfo{journal}{IEEE Transactions on Neural Systems and
  Rehabilitation Engineering}\/},  {\it \bibinfo{volume}{11}\/},
  \bibinfo{pages}{141--144}.
\bibitem[{Guyon \& Elisseeff(2003)}]{Guyon:2003}
\bibinfo{author}{Guyon, I.}, \& \bibinfo{author}{Elisseeff, A.}
  (\bibinfo{year}{2003}).
\newblock \bibinfo{title}{An introduction to variable and feature selection}.
\newblock {\it \bibinfo{journal}{Journal of Machine Learning Research}\/},
  {\it \bibinfo{volume}{3}\/}, \bibinfo{pages}{1157--1182}.
\bibitem[{Guyon et~al.(2002)Guyon, Weston, Barnhill \& Vapnik}]{Guyon:2002}
\bibinfo{author}{Guyon, I.}, \bibinfo{author}{Weston, J.},
  \bibinfo{author}{Barnhill, S.}, \& \bibinfo{author}{Vapnik, V.}
  (\bibinfo{year}{2002}).
\newblock \bibinfo{title}{Gene selection for cancer classification using
  support vector machines}.
\newblock {\it \bibinfo{journal}{Machine Learning}\/},  {\it
  \bibinfo{volume}{46}\/}, \bibinfo{pages}{389--422}.
\bibitem[{Hellman \& Raviv(1970)}]{Hellman:1970}
\bibinfo{author}{Hellman, M.}, \& \bibinfo{author}{Raviv, J.}
  (\bibinfo{year}{1970}).
\newblock \bibinfo{title}{Probability of error, equivocation, and the
  \uppercase{C}hernoff bound}.
\newblock {\it \bibinfo{journal}{IEEE Transactions on Information Theory}\/},
  {\it \bibinfo{volume}{16}\/}, \bibinfo{pages}{368--372}.
\bibitem[{Hild et~al.(2006)Hild, Erdogmus, Torkkola \& Principe}]{Hild:2006}
\bibinfo{author}{Hild, K.~E.}, \bibinfo{author}{Erdogmus, D.},
  \bibinfo{author}{Torkkola, K.}, \& \bibinfo{author}{Principe, J.~C.}
  (\bibinfo{year}{2006}).
\newblock \bibinfo{title}{Feature extraction using information--theoretic
  learning}.
\newblock {\it \bibinfo{journal}{IEEE Transactions on Pattern Analysis and
  Machine Intelligence}\/},  {\it \bibinfo{volume}{28}\/},
  \bibinfo{pages}{1385--1392}.
\bibitem[{Hinton \& Salakhutdinov(2006)}]{Hinton:2006}
\bibinfo{author}{Hinton, G.~E.}, \& \bibinfo{author}{Salakhutdinov, R.~R.}
  (\bibinfo{year}{2006}).
\newblock \bibinfo{title}{Reducing the dimensionality of data with neural
  networks}.
\newblock {\it \bibinfo{journal}{Science}\/},  {\it \bibinfo{volume}{313}\/},
  \bibinfo{pages}{504--507}.
\bibitem[{Hjelm et~al.(2018)Hjelm, Fedorov, Lavoie-Marchildon, Grewal, Bachman,
  Trischler \& Bengio}]{Hjelm:2018}
\bibinfo{author}{Hjelm, R.~D.}, \bibinfo{author}{Fedorov, A.},
  \bibinfo{author}{Lavoie-Marchildon, S.}, \bibinfo{author}{Grewal, K.},
  \bibinfo{author}{Bachman, P.}, \bibinfo{author}{Trischler, A.}, \&
  \bibinfo{author}{Bengio, Y.} (\bibinfo{year}{2018}).
\newblock \bibinfo{title}{Learning deep representations by mutual information
  estimation and maximization}.
\newblock {\it \bibinfo{journal}{arXiv preprint arXiv:1808.06670}\/}, .
\bibitem[{Hwang et~al.(1994)Hwang, Lay \& Lippman}]{Hwang:1994}
\bibinfo{author}{Hwang, J.-N.}, \bibinfo{author}{Lay, S.-R.}, \&
  \bibinfo{author}{Lippman, A.} (\bibinfo{year}{1994}).
\newblock \bibinfo{title}{Nonparametric multivariate density estimation: a
  comparative study}.
\newblock {\it \bibinfo{journal}{IEEE Transactions on Signal Processing}\/},
  {\it \bibinfo{volume}{42}\/}, \bibinfo{pages}{2795--2810}.
\bibitem[{Jacob et~al.(2009)Jacob, Obozinski \& Vert}]{Jacob:2009}
\bibinfo{author}{Jacob, L.}, \bibinfo{author}{Obozinski, G.}, \&
  \bibinfo{author}{Vert, J.-P.} (\bibinfo{year}{2009}).
\newblock \bibinfo{title}{Group lasso with overlap and graph lasso}.
\newblock In {\it \bibinfo{booktitle}{Proceedings of the 26th Annual
  International Conference on Machine Learning}\/} (pp.
  \bibinfo{pages}{433--440}).
\newblock \bibinfo{organization}{ACM}.
\bibitem[{Jiang et~al.(2016)Jiang, Zhang, Ren, Han, Chen \& Hanzo}]{Jiang:2016}
\bibinfo{author}{Jiang, C.}, \bibinfo{author}{Zhang, H.}, \bibinfo{author}{Ren,
  Y.}, \bibinfo{author}{Han, Z.}, \bibinfo{author}{Chen, K.-C.}, \&
  \bibinfo{author}{Hanzo, L.} (\bibinfo{year}{2016}).
\newblock \bibinfo{title}{Machine learning paradigms for next-generation
  wireless networks}.
\newblock {\it \bibinfo{journal}{IEEE Wireless Communications}\/},  {\it
  \bibinfo{volume}{24}\/}, \bibinfo{pages}{98--105}.
\bibitem[{Kwak \& Choi(2002)}]{Kwak:2002}
\bibinfo{author}{Kwak, N.}, \& \bibinfo{author}{Choi, C.~H.}
  (\bibinfo{year}{2002}).
\newblock \bibinfo{title}{Input feature selection by mutual information based
  on parzen window}.
\newblock {\it \bibinfo{journal}{IEEE Transactions on Pattern Analysis and
  Machine Intelligence}\/},  {\it \bibinfo{volume}{24}\/},
  \bibinfo{pages}{1667--1671}.
\bibitem[{Larranaga et~al.(2006)}]{Larranaga:2006}
\bibinfo{author}{Larranaga, P.} et~al. (\bibinfo{year}{2006}).
\newblock \bibinfo{title}{Machine learning in bioinformatics}.
\newblock {\it \bibinfo{journal}{Briefings in Bioinformatics}\/},  {\it
  \bibinfo{volume}{7}\/}, \bibinfo{pages}{86--112}.
\bibitem[{Lazar et~al.(2012)Lazar, Taminau, Meganck, Steenhoff, Coletta,
  Molter, de~Schaetzen, Duque, Bersini \& Nowe}]{Lazar:2012}
\bibinfo{author}{Lazar, C.}, \bibinfo{author}{Taminau, J.},
  \bibinfo{author}{Meganck, S.}, \bibinfo{author}{Steenhoff, D.},
  \bibinfo{author}{Coletta, A.}, \bibinfo{author}{Molter, C.},
  \bibinfo{author}{de~Schaetzen, V.}, \bibinfo{author}{Duque, R.},
  \bibinfo{author}{Bersini, H.}, \& \bibinfo{author}{Nowe, A.}
  (\bibinfo{year}{2012}).
\newblock \bibinfo{title}{A survey on filter techniques for feature selection
  in gene expression microarray analysis}.
\newblock {\it \bibinfo{journal}{IEEE/ACM Transactions on Computational Biology
  and Bioinformatics}\/},  {\it \bibinfo{volume}{9}\/},
  \bibinfo{pages}{1106--1119}.
\bibitem[{Lemm et~al.(2011)Lemm, Blankertz, Dickhaus \& M{\"u}ller}]{Lemm:2011}
\bibinfo{author}{Lemm, S.}, \bibinfo{author}{Blankertz, B.},
  \bibinfo{author}{Dickhaus, T.}, \& \bibinfo{author}{M{\"u}ller, K.-R.}
  (\bibinfo{year}{2011}).
\newblock \bibinfo{title}{Introduction to machine learning for brain imaging}.
\newblock {\it \bibinfo{journal}{NeuroImage}\/},  {\it \bibinfo{volume}{56}\/},
  \bibinfo{pages}{387--399}.
\bibitem[{Montavon et~al.(2018)Montavon, Samek \& M{\"u}ller}]{Montavon:2018}
\bibinfo{author}{Montavon, G.}, \bibinfo{author}{Samek, W.}, \&
  \bibinfo{author}{M{\"u}ller, K.-R.} (\bibinfo{year}{2018}).
\newblock \bibinfo{title}{Methods for interpreting and understanding deep
  neural networks}.
\newblock {\it \bibinfo{journal}{Digital Signal Processing}\/},  {\it
  \bibinfo{volume}{73}\/}, \bibinfo{pages}{1--15}.
\bibitem[{Nenadic(2007)}]{Nenadic:2007}
\bibinfo{author}{Nenadic, Z.} (\bibinfo{year}{2007}).
\newblock \bibinfo{title}{Information discriminant analysis: Feature extraction
  with an information-theoretic objective}.
\newblock {\it \bibinfo{journal}{IEEE Transactions on Pattern Analysis and
  Machine Intelligence}\/},  {\it \bibinfo{volume}{29}\/},
  \bibinfo{pages}{1394--1407}.
\bibitem[{Nutt et~al.(2003)}]{Nutt:2003}
\bibinfo{author}{Nutt, C.~L.} et~al. (\bibinfo{year}{2003}).
\newblock \bibinfo{title}{Gene expression-based classification of malignant
  gliomas correlates better with survival than histological classification}.
\newblock {\it \bibinfo{journal}{Cancer research}\/},  {\it
  \bibinfo{volume}{63}\/}, \bibinfo{pages}{1602--1607}.
\bibitem[{\"{O}zdenizci \& Erdo\u{g}mu\c{s}(2020)}]{Ozdenizci:2019}
\bibinfo{author}{\"{O}zdenizci, O.}, \& \bibinfo{author}{Erdo\u{g}mu\c{s}, D.}
  (\bibinfo{year}{2020}).
\newblock \bibinfo{title}{Information theoretic feature transformation learning
  for brain interfaces}.
\newblock {\it \bibinfo{journal}{IEEE Transactions on Biomedical
  Engineering}\/},  {\it \bibinfo{volume}{67}\/}, \bibinfo{pages}{69--78}.
\bibitem[{{\"O}zdenizci et~al.(2020){\"O}zdenizci, Wang, Koike-Akino \&
  Erdo{\u{g}}mu{\c{s}}}]{Ozdenizci:2020}
\bibinfo{author}{{\"O}zdenizci, O.}, \bibinfo{author}{Wang, Y.},
  \bibinfo{author}{Koike-Akino, T.}, \& \bibinfo{author}{Erdo{\u{g}}mu{\c{s}},
  D.} (\bibinfo{year}{2020}).
\newblock \bibinfo{title}{Learning invariant representations from
  \uppercase{EEG} via adversarial inference}.
\newblock {\it \bibinfo{journal}{IEEE Access}\/},  {\it \bibinfo{volume}{8}\/},
  \bibinfo{pages}{27074--27085}.
\bibitem[{Parzen(1962)}]{Parzen:1962}
\bibinfo{author}{Parzen, E.} (\bibinfo{year}{1962}).
\newblock \bibinfo{title}{On estimation of a probability density function and
  mode}.
\newblock {\it \bibinfo{journal}{The Annals of Mathematical Statistics}\/},
  {\it \bibinfo{volume}{33}\/}, \bibinfo{pages}{1065--1076}.
\bibitem[{Peng et~al.(2005)Peng, Long \& Ding}]{Peng:2005}
\bibinfo{author}{Peng, H.}, \bibinfo{author}{Long, F.}, \&
  \bibinfo{author}{Ding, C.} (\bibinfo{year}{2005}).
\newblock \bibinfo{title}{Feature selection based on mutual information
  criteria of max-dependency, max-relevance, and min-redundancy}.
\newblock {\it \bibinfo{journal}{IEEE Transactions on Pattern Analysis and
  Machine Intelligence}\/},  {\it \bibinfo{volume}{27}\/},
  \bibinfo{pages}{1226--1238}.
\bibitem[{Principe et~al.(2000)Principe, Xu, Fisher \& Haykin}]{Principe:2000}
\bibinfo{author}{Principe, J.~C.}, \bibinfo{author}{Xu, D.},
  \bibinfo{author}{Fisher, J.}, \& \bibinfo{author}{Haykin, S.}
  (\bibinfo{year}{2000}).
\newblock \bibinfo{title}{Information theoretic learning}.
\newblock {\it \bibinfo{journal}{Unsupervised Adaptive Filtering}\/},  {\it
  \bibinfo{volume}{1}\/}, \bibinfo{pages}{265--319}.
\bibitem[{Qian(1999)}]{Qian:1999}
\bibinfo{author}{Qian, N.} (\bibinfo{year}{1999}).
\newblock \bibinfo{title}{On the momentum term in gradient descent learning
  algorithms}.
\newblock {\it \bibinfo{journal}{Neural Networks}\/},  {\it
  \bibinfo{volume}{12}\/}, \bibinfo{pages}{145--151}.
\bibitem[{Ross(2014)}]{Ross:2014}
\bibinfo{author}{Ross, B.~C.} (\bibinfo{year}{2014}).
\newblock \bibinfo{title}{Mutual information between discrete and continuous
  data sets}.
\newblock {\it \bibinfo{journal}{PloS one}\/},  {\it \bibinfo{volume}{9}\/},
  \bibinfo{pages}{e87357}.
\bibitem[{Sanchez et~al.(2019)Sanchez, Serrurier \& Ortner}]{Sanchez:2019}
\bibinfo{author}{Sanchez, E.~H.}, \bibinfo{author}{Serrurier, M.}, \&
  \bibinfo{author}{Ortner, M.} (\bibinfo{year}{2019}).
\newblock \bibinfo{title}{Learning disentangled representations via mutual
  information estimation}.
\newblock {\it \bibinfo{journal}{arXiv preprint arXiv:1912.03915}\/}, .
\bibitem[{Silverman(1986)}]{Silverman:1986}
\bibinfo{author}{Silverman, B.~W.} (\bibinfo{year}{1986}).
\newblock {\it \bibinfo{title}{Density Estimation for Statistics and Data
  Analysis}\/}.
\newblock \bibinfo{publisher}{Chapman \& Hall}.
\bibitem[{Thrun et~al.(1991)}]{Thrun:1991}
\bibinfo{author}{Thrun, S.~B.} et~al. (\bibinfo{year}{1991}).
\newblock \bibinfo{title}{The \uppercase{MONK}'s problems a performance
  comparison of different learning algorithms}, .
\bibitem[{Tokui et~al.(2015)Tokui, Oono, Hido \& Clayton}]{Tokui:2015}
\bibinfo{author}{Tokui, S.}, \bibinfo{author}{Oono, K.}, \bibinfo{author}{Hido,
  S.}, \& \bibinfo{author}{Clayton, J.} (\bibinfo{year}{2015}).
\newblock \bibinfo{title}{Chainer: a next-generation open source framework for
  deep learning}.
\newblock In {\it \bibinfo{booktitle}{Proceedings of Workshop on Machine
  Learning Systems in the Twenty-ninth Annual Conference on Neural Information
  Processing Systems}\/} (pp. \bibinfo{pages}{1--6}).
\newblock volume~\bibinfo{volume}{5}.
\bibitem[{Torkkola(2003)}]{Torkkola:2003}
\bibinfo{author}{Torkkola, K.} (\bibinfo{year}{2003}).
\newblock \bibinfo{title}{Feature extraction by non-parametric mutual
  information maximization}.
\newblock {\it \bibinfo{journal}{Journal of Machine Learning Research}\/},
  {\it \bibinfo{volume}{3}\/}, \bibinfo{pages}{1415--1438}.
\bibitem[{Torkkola(2008)}]{Torkkola:2008}
\bibinfo{author}{Torkkola, K.} (\bibinfo{year}{2008}).
\newblock \bibinfo{title}{Information-theoretic methods}.
\newblock In {\it \bibinfo{booktitle}{Feature Extraction}\/} (pp.
  \bibinfo{pages}{167--185}).
\newblock \bibinfo{publisher}{Springer}.
\bibitem[{Wen et~al.(2020)Wen, Zhou, He, Zhou \& Xu}]{Wen:2020}
\bibinfo{author}{Wen, L.}, \bibinfo{author}{Zhou, Y.}, \bibinfo{author}{He,
  L.}, \bibinfo{author}{Zhou, M.}, \& \bibinfo{author}{Xu, Z.}
  (\bibinfo{year}{2020}).
\newblock \bibinfo{title}{Mutual information gradient estimation for
  representation learning}.
\newblock {\it \bibinfo{journal}{arXiv preprint arXiv:2005.01123}\/}, .
\bibitem[{Xia et~al.(2020)Xia, Zhou, Shi, Lu \& Huang}]{Xia:2020}
\bibinfo{author}{Xia, Y.}, \bibinfo{author}{Zhou, J.}, \bibinfo{author}{Shi,
  Z.}, \bibinfo{author}{Lu, C.}, \& \bibinfo{author}{Huang, H.}
  (\bibinfo{year}{2020}).
\newblock \bibinfo{title}{Generative adversarial regularized mutual information
  policy gradient framework for automatic diagnosis}.
\newblock In {\it \bibinfo{booktitle}{Proceedings of the AAAI Conference on
  Artificial Intelligence}\/} (pp. \bibinfo{pages}{1062--1069}).
\newblock volume~\bibinfo{volume}{34}.
\bibitem[{Zhang et~al.(2010)Zhang, Guan \& Ang}]{Zhang:2010}
\bibinfo{author}{Zhang, H.}, \bibinfo{author}{Guan, C.}, \&
  \bibinfo{author}{Ang, K.~K.} (\bibinfo{year}{2010}).
\newblock \bibinfo{title}{An information theoretic linear discriminant analysis
  method}.
\newblock In {\it \bibinfo{booktitle}{International Conference on Pattern
  Recognition}\/} (pp. \bibinfo{pages}{4182--4185}).
\bibitem[{Zhu et~al.(2004)Zhu, Rosset, Tibshirani \& Hastie}]{Zhu:2004}
\bibinfo{author}{Zhu, J.}, \bibinfo{author}{Rosset, S.},
  \bibinfo{author}{Tibshirani, R.}, \& \bibinfo{author}{Hastie, T.~J.}
  (\bibinfo{year}{2004}).
\newblock \bibinfo{title}{1-norm support vector machines}.
\newblock In {\it \bibinfo{booktitle}{Advances in Neural Information Processing
  Systems}\/} (pp. \bibinfo{pages}{49--56}).

\end{thebibliography}

\end{document}